# Using Large Language Models to Automate Category and Trend Analysis of Scientific Articles: An Application in Ophthalmology

Hina Raja, Asim Munawar, Mohammad Delsoz, Mohammad Elahi, Yeganeh Madadi, Amr Hassan, Hashem Abu Serhan, Onur Inam, Luis Hermandez, Sang Tran, Wuqas Munir, Alaa Abd-Alrazaq, Hao Chen, and SiamakYousefi

*Abstract*— **Purpose: In this paper, we present an automated method for article classification, leveraging the power of Large Language Models (LLM). The primary focus is on the field of ophthalmology, but the model is extendable to other fields. Methods: We have developed a model based on Natural Language Processing (NLP) techniques, including advanced LLMs, to process and analyze the textual content of scientific papers. Specifically, we have employed zero-shot learning (ZSL) LLM models and compared against Bidirectional and Auto-Regressive Transformers (BART) and its variants, and Bidirectional Encoder Representations from Transformers (BERT), and its variant such as distilBERT, SciBERT, PubmedBERT, BioBERT. Results: The classification results demonstrate the effectiveness of LLMs in categorizing large number of ophthalmology papers without human intervention. Results: To evalute the LLMs, we compiled a dataset (RenD) of 1000 ocular disease-related articles, which were expertly annotated by a panel of six specialists into 15 distinct categories. The model achieved mean accuracy of 0.86 and mean F1 of 0.85 based on the RenD dataset. Conclusion: The proposed framework achieves notable improvements in both accuracy and efficiency. Its application in the domain of ophthalmology showcases its potential for knowledge organization and retrieval in other domains too. We performed trend analysis that enables the researchers and clinicians to easily categorize and retrieve relevant papers, saving time and effort in literature review and information gathering as well as identification of emerging scientific trends within different disciplines. Moreover, the extendibility of the model to other scientific fields broadens its impact in facilitating research and trend analysis across diverse disciplines.**

## 1. INTRODUCTION

A literature review is an integral component of the research process that involves systematically reviewing, evaluating, and synthesizing existing scholarly publications from databases such as Medline/PubMed, Embase, and Google scholar. The standard approach for literature review involves using a bibliographic search engine to conduct an initial comprehensive search. Researchers utilize relevant keywords and filters, including clinical query filters, to retrieve a wide range of articles. Next steps include manually screening the retrieved articles by reviewing titles, abstracts, and, in most cases, full texts to assess their relevance and inclusion criteria. This combination of automated search and manual screening ensures a thorough review while targeting specific research objectives.

However, literature review can be a challenging and time-consuming task for researchers, requiring meticulous examination of numerous sources and critical analysis of their findings. The process demands substantial time and effort to effectively navigate through the vast expanse of scholarly literature from different databases and extract meaningful insights. Artificial Intelligence tools have been used for facilitating this search process  [1].

Hina Raja. Hamilton Eye Institute, Department of Ophthalmology, University of Tennessee Health Science Center, Memphis, TN, US (hraja@uthsc.edu ).

Asim Munawar. Program Director for Neuro-Symbolic AI at IBM Research, USA (asim@ibm.com )

Nikolaos Mylonas. Aristotle University of Thessaloniki (AUTH), Thessaloniki, 54124, Greece. (myloniko@csd.auth.gr )

Mohammad Delsoz. Hamilton Eye Institute, Department of Ophthalmology, University of Tennessee Health Science Center, Memphis, TN, US (mdelsoz@uthsc.edu).

Yeganeh Mehdi. Hamilton Eye Institute, Department of Ophthalmology, University of Tennessee Health Science Center, Memphis, TN, US (ymehdi@uthsc.edu ).

Mohammad Elahi. Quillen College of Medicine, East Tennessee State University, TN, USA. (m.elahi047@gmail.com).

Amr Hassan. Department of Ophthalmology, Faculty of Medicine, South Valley University, Egypt (dramrhassan@yahoo.com ).

Hashem Abu Serhan. Department of Ophthalmology, Hamad Medical Corporation, Doha, Qatar (hashemabusarhan@yahoo.com).

Onur Inam,  1.Department of Biophysics, Faculty of Medicine, Gazi University,Turkey 2. Department of Ophthalmology, Edward S. Harkness Eye Institute, Vagelos College of Physicians and Surgeons, Columbia University Irving Medical Center, New York, NY, USA (onurinam1990@hotmail.com).

Luis Hermandez. Association to Prevent Blindness in Mexico (lahpm092@gmail.com).

Hao Chen, Department of Pharmacology, Addiction Science and Toxicology, and Department of Ophthalmology, University of Tennessee Health Science Center, Memphis, TN, US (hchen3@uthsc.edu)

Sang Tran, University of Maryland, School of Medicine, Department of Ophthalmology and Visual Sciences, Baltimore, MD 21201 Add affiliation (sang.tran@som.umaryland.edu)

Wuqas Munir, University of Maryland School of Medicine, (wmunir@som.umaryland.edu)

Alaa Abd-Atraz, AI Center for Precision Health, Weill Cornell Medicine-Qatar, Doha, Qatar (aaa4027@qatar-med.cornell.edu).

Siamak Yousefi. Hamilton Eye Institute, Department of Ophthalmology, University of Tennessee Health Science Center, Memphis, TN, US (siamak.yousefi@uthsc.edu ).

Machine learning models have been applied to perform the text classification task based on feature engineering [2-4]. A semi-automated model was proposed for article classification in systemic review articles based on mechanistic pathways [5]. A total of 24,737 abstracts from PubMed and Web of Science databases and 861 references were found to be relevant. They evaluated the Naïve Bayes, Support Vector Machines (SVM), regularized logistic regressions, neural networks, random forest, Logit boost, and XGBoost models. The best performing model achieved sensitivity and specificity of ~70% and ~60%, respectively. Kanegasaki et al. [6] used Long Short-Term Memory networks (LSTM) for classification of abstracts. They used two datasets with 1307 and 1023 articles and achieved 73% and 77% in correctly classifying the abstracts based on two the datasets, respectively.

Natural Language Processing (NLP) applications have significantly advanced in recent years and gained tremendous popularity due to their wide range of applications across various domains. With the increasing availability of large datasets and advancements in computational power, NLP has made remarkable progress, revolutionizing the way we interact with technology [7-10]. Particularly, NLP has gained interest in the field of information retrieval . NLP techniques, such as keyword extraction, document clustering, and semantic search, have improved the accuracy and relevance of search results.

Domocos et al. [11] used the Bert model for classifying articles into human, animal and in-vivo groups. Ashwin et al. [12] employed SciBERT to classify scientific articles in four major categories including format, human health care (HHC), purpose, and rigor. The format category included original study, review, case report, and general articles. HHC encompassed all articles discussing human health. The purpose category included articles discussing etiology, diagnosis, prognosis, treatment, costs, economics, and disease-related prediction. Rigor class included the studies that presented design criteria specific to a class purpose. The model achieved F1 score of 0.753 on the publicly available Clinical Hedges dataset.

Kim et al. [13] used the BERT model for classification of scientific articles on randomized controlled trials (RCTs). The BioBERT variant, trained on titles and abstracts, showed the highest performance of 0.90 in terms of F1 score. Another study [14] fine-tuned variants of BERT model including BERTBASE, BlueBERT, PubMedBERT, and BioBERT for classification of human health studies. They have used abstracts and titles of 160,000 articles from the PubMed database. BioBERT showed the best results and achieved 60%-70% specificity and recall >90%. The study [15] proposed weakly supervised classification of biomedical articles. The model was trained on weakly labeled subset of the BioASQ 2018 dataset based on Mesh descriptors. BioBERT was used to generate the embedding for words and sentences and then utilized the cosine similarity to assign labels. The proposed model achieved F1 score of 0.564 on the BioASQ 2020 dataset.

Conventional approaches to text classification have traditionally relied on the assumption that there is a fixed set of predefined labels to which a given article can be assigned. However, this assumption is violated when dealing with real-world applications, where the label space for describing a text is virtually unlimited, and the potential labels that can be associated with a text span an infinite spectrum, reflecting the diverse and nuanced nature of textual content. Such complexity challenges the conventional methods and calls for innovative strategies to navigate the expansive and unbounded label space. To address these issues zero shot techniques [16-18] are developed and are gaining popularity. Zero-shot learning (ZSL) involves classifying instances into categories without any labeled training data [19]. It leverages auxiliary information like semantic embeddings or textual descriptions to bridge the gap between known and unknown categories. This enables models to generalize to novel classes and make predictions for unseen categories. Mylonas et al. [20] employed zero-shot classification model for PubMed articles into emerging MESH descriptors. Instead of using the standard n-grams approach, the method exploited BioBERT embeddings at the sentence level to turn textual input into a new semantic space for the clinical Hedges dataset [21].

In this work we have employed Large Language Models (LLM) models that include ZSL for categorizing the ophthalmology articles from the PubMed database into different categories based on title and abstract. We fine-tuned the BERT and its variants BERTBASE, SciBERT, PubmedBERT, and BioBERT for those categories which showed not comprising results from ZSL model. In addition , we performed trend analysis based on the classified results, providing researchers insights into emerging trends in the field to stay updated on the latest developments and identifying key areas of interest. Overall, we provide a method that enhances the efficiency, relevance, and interdisciplinary potential of the literature review process.

The rest of the paper is organized as follows: Section 2 presents materials and methods of the proposed framework for the text classification and trend analysis. Whereas Section 3 discusses the results of different experiments performed for the evaluation of the proposed model. Section 4 presents the discussion and, Section 5 concludes the proposed work.

**What is Already Known** Various classification models have been proposed in the literature for biomedical articles to retrieve relevant information [1-21]. Fine-tuned BERT and its variants have been used for text classification.

**What this Paper Add** We have explored the ZSL models for classification of biomedical articles. We have developed different use cases targeting the field of ophthalmology. To evaluate the model, we have generated a dataset that includes 1000 articles related to ocular diseases. Articles were manually annotated by six experts into 15 categories. ZSL model BART achieved mean accuracy of 0.86 and F1 score of 0.85. In addition to classification of articles, we also performed trend analysis on different classified groups. The model is adaptable to other biomedical disciplines without explicit fine-tuning and training.

## 2. Materials and Methods

### 2.1 Dataset

There are several annotated datasets available for various NLP tasks in the biomedical domain. However in the field of ophthalmology, there is a scarcity of publicly accessible datasets for performing NLP tasks. To address this gap, we have taken the initiative to curate a dataset focused on ocular diseases. Our retinal diseases (RenD) dataset comprises of 1000 articles sourced from PubMed, covering various conditions such as diabetic retinopathy (DR), glaucoma, diabetic macular edema (DME), age-related macular degeneration (AMD), cataract, dry eye, retinal detachment, and central serous retinopathy (CSR). To ensure accurate categorization, we enlisted the expertise of six domain specialists who meticulously annotated the articles based on abstracts. To ensure accuracy and reliability in the annotation process, each article in our dataset is reviewed and annotated by at least three individual annotators. This multiple-annotator approach helps mitigate potential biases and inconsistencies that could arise from a single annotator's perspective. Once the annotation was completed, the final label for each article is determined based on majority voting. This dataset, with its 15 distinct labels (see supplemental Table 1 for guidelines for the data annotation) and grouped into four categories. We will make this dataset publicly accessible to the community for advancing research, facilitating comprehensive analysis, enabling more targeted investigations into ocular diseases, and to promoting open science.

## 2.2 Models

Our framework automates the entire literature review process in the ophthalmology domain (Figure 1). More specifically, by incorporating user-defined criteria, including keyword, number of articles, inclusion criteria, and categories for classification, our framework performs a systematic retrieval and analysis of relevant articles automatically. Based on the keyword model extracts the abstracts, titles, publication years, and links from PubMed database. Preprocessing is then performed based on the inclusion criteria, followed by article classification. To improve accuracy, ZSL was leveraged initially, followed by BERT fine-tuning for categories that ZSL exhibits limitations.

## 2.3 Zero-shot Classification

Zero-shot classification is an approach to predict the class of instances for categories they have never seen during training, using auxiliary information or semantic embeddings. It enables generalization to unseen classes and expands the classification capabilities beyond the limitations of labeled training data. We have employed Bidirectional and Auto-Regressive Transformers (BART) [22] pretrained based on sequence-to-sequence model that combines bidirectional and auto-regressive techniques for improved text generation and comprehension. BART has 12 transformer layers with hidden size of 1024 that was initially trained on Wikipedia and BookCorpus dataset and fine-tuned on Multi-Genre Natural Language Inference (MNLI) tasks. BART is an amalgamation of the bi-directional encoder found in BERT and the autoregressive decoder used in GPT. While BERT comprises approximately 110 million trainable parameters and GPT-3 consists of 117 million parameters, BART, being a combination of the two, has approximately 140-400 million parameters. This larger parameter count in BART accommodates its sequenced structure, which incorporates both encoding and decoding capabilities for a wide range of NLP tasks. The model receives the title and abstract of an article as input and generates probabilities for different categories. The final label for multiclass classification of article is determined by taking the maximum probability among all classes (Eq 1), and for multilabel classification class labels are assigned as probability is greater than threshold value (Eq 2). In these equations, $p(i)$ is the probability of article based on title and abstract, C is the total number of classes for particular category, and $\xi$ is threshold.

$$L_{fin\_MC} = max|(p(i))|\;_{i=1}^{C} \qquad (1)$$

$$L_{fin\_ML} = |(p(i))|\;_{i=1}^{C} > \xi \qquad (2)$$

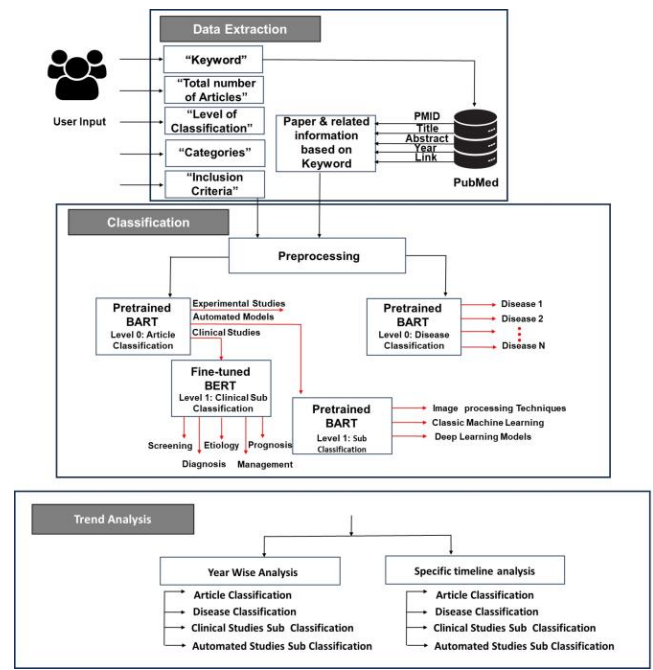

Fig. 1. Flow diagram of proposed framework.

## 2.4 Multilabel Classification

BERT [13] and its variant models, namely distilBERT, SciBERT, PubmedBERT, and BioBERT, have been subjected to fine-tuning to address multilabel classification tasks for categories where ZSL model (BART) is unable to produce more accurate results. The preprocessing stage entails the concatenation of article titles and abstracts, which are subsequently input into the respective BERT model. The model generates probabilities for each class, and if the probability for a specific category is higher than a threshold value, the article is assigned the label corresponding to that category.

## 2.5. Trend Analysis

In addition to classification of articles, we performed two additional analyses as well. More specifically, we performed a technology trend analysis to obtain valuable insights into the distribution of research across different classes, highlighting classes with higher or lower publication frequencies. This information aids in understanding the emphasis and focus of research efforts, enabling resource allocation and identifying areas that may require further attention or investigation. We also performed an interest trend analysis to provide a comprehensive view of publication trends over specific time periods. By identifying the popularity of techniques or topics over time, this analysis facilitates the detection of emerging trends and the evaluation of long-term patterns. These trend analysis, applicable to all levels of classification categories, contribute to an enhanced understanding of the dynamic nature and evolving landscape of research in the field.

TABLE 1

DATASETS DESCRIPTION, $N_T$ IS THE TOTAL NUMBER OF ARTICLES IN EACH DATASET, $N_C$ REPRESENTS THE TOTAL NUMBER OF CATEGORIES IN EACH DATASET.

| Dataset | $N_T$ | $N_C$ | Group | Categories |
|---|---|---|---|---|
| **Retinal disease (RenD)** | 1000 | 19 | Article Type | Clinical, Experimental, and Automated Model |
| | | | Ocular Diseases | DR, DME, AMD, glaucoma, Dry Eye, Cataract, CSR, and retinal detachment |
| | | | Clinical Studies Sub-Class | Screening, Diagnosis, Prognosis, Etiology, and Management |
| | | | Automated Studies Sub-Class | Image processing techniques, Machine learning models, and Deep learning models |
| **Dry Eye (DEye)** | 67 | 6 | Clinical Studies Sub-Class | Tear Film Break Up Time, Infrared Thermography, Lipid Layer Interface Pattern, Meibomian Gland Study, Blink Study, Tear Film Assessment, Tear Meniscus Assessment |
| **Glaucoma (DemL)** | 115 | 2 | Automated Studies Sub-Class | Machine learning model, Deep learning model |

## 3. RESULTS

This section presents the experiments we have performed to evaluate the LLM models for classification.

### 3.1 Experimental Details

We evaluated our models based on the dataset retinal disease (RenD) dataset that we annotated. Additionally, we evaluated the models to classify categories based on two review studies related to dry eye disease and glaucoma (see Table 1). We present results in terms of accuracy (Ac), Area Under the Curve (AUC), F1 score (F1), precision (Pv) and recall (Re) for each dataset. For multilabel classification, we evaluated the model in terms of F1 micro, Pv micro, Re micro, and AUC.

### 3.2 Ablation Study

Our research encompasses both multiclass and multilabel classification tasks. To accomplish this, we employed ZSL model and fine-tuned the model for which ZSL was not performing well. Through a series of ablation experiments, we systematically investigated the impact of different settings (see sub sections) on the performance of the model. By modifying and assessing various settings, we gained insights into the individual contributions and effects of each setting, allowing us to refine and optimize our approach accordingly.

#### 3.2.1 ZSL Model Selection

In our study, we conducted an evaluation of the ZSL state-of-the-art models and multiple variants of the BART model. Based on a comprehensive analysis, we identified the model variant that exhibited the most favorable performance in our specific context (see Table 2). We selected BART model that showed the best performance. Additionally, we performed experiments by employing different keywords for the different categories. It was observed that for ZSL model, the keywords should be more descriptive and provide some information about related categories to improve the accuracy (see Table 3). For category "Clinical, Experimental, and Automated Model", we have tested various keywords and found that 'Clinical finding based on humans', 'Experimental study based on animals', 'Technical study based on automated model' keywords showed the best results with accuracy of 0.91 and F1 score of 0.92 . During our evaluation, we investigated the potential of using abstracts and titles for classification across various categories. Surprisingly, we discovered that classification solely based on titles closely approximates the results obtained from using abstracts for most of the categories. However, abstract and title both together enhances the efficacy of classification.

#### 3.2.2 Hyperparameter Tuning

For the categories where ZSL model (BART) provided poor results, we fine-tuned the BERT model and its variants to perform categorization. We conducted hyperparameter tuning based to enhance the model's reliability and significance. By carefully selecting and fine-tuning hyperparameters such as learning rates, batch sizes, and regularization strengths, we aimed to achieve accurate and meaningful results (see supplementary Table A2). For the BioBERT model, we selected a learning rate of 1e-05, batch size of 8, max length of 400, and number of epochs of 20.

### 3.3 Evaluation Results

#### 3.3.1. Article Classification Evaluation:

This section presents the results, based on the metrics which were selected in the ablation experiments. Based on the evaluation, BART demonstrated the best performance among the tested models. As a result, further classification tasks were conducted using the BART model to capitalize on its superior performance. Table 4 shows the classification results using BART for RenD dataset for categories article type, ocular diseases, clinical studies sub-class, and automated studies sub-class. Article type group was classified into three categories: clinical, experiment and automated studies. The BART model demonstrated impressive performance for article type group, with an accuracy of 0.91, an F1 score of 0.92, an AUC of 0.91, precision of 0.93, and recall of 0.91. For article group type, abstract and title and both are performing consistent. Whereas automated category is further categories into image processing techniques, machine learning models, and deep learning models. For automated study sub-class group, ZSL model achieved best performance for title-based classification with accuracy, F1 score, AUC, precision, and recall of 0.92, 0.92,

0.95, 0.94, and 0.92, respectively. However, second best scores were achieved by classification based on abstract and title.

Clinical studies are further categorized into screening, diagnosis, prognosis, etiology, and management, constituting a

TABLE 2

Evaluation of ZSL classification models for the category 1 from RenD dataset.

| Models | Time (min) | Abstract | | | | | Title | | | | |
|---|---|---|---|---|---|---|---|---|---|---|---|
| | | Ac | F1 | AUC | Pv | Re | Ac | F1 | AUC | Pv | Re |
| **Bart-base** | 34.34 3.5 | 0.08 | 0.03 | 0.42 | 0.86 | 0.08 | 0.01 | 0.005 | 0.5 | 0.006 | 0.01 |
| **Bart-large** | 12.24 | 0.11 | 0.03 | 0.46 | 0.17 | 0.115 | 0.5 | 0.57 | 0.39 | 0.68 | 0.50 |
| **Bart-large-cnn** | 37.63 | 0.08 | 0.03 | 0.42 | 0.86 | 0.08 | 0.36 | 0.50 | 0.58 | 0.84 | 0.36 |
| **Bart-mnli-CNN** | 231.42 17.76 | 0.74 | 0.78 | 0.65 | 0.84 | 0.74 | 0.09 | 0.06 | 0.42 | 0.2 | 0.09 |
| **Mdeberta-v3-base** | 79.28 31.06 | 0.87 | 0.85 | 0.74 | 0.88 | 0.87 | 0.76 | 0.82 | 0.80 | 0.91 | 0.76 |
| **Bart-large-mnli** | **141.50 15.21** | **0.91** | **0.92** | **0.91** | **0.93** | **0.91** | **0.91** | **0.82** | **0.93** | **0.94** | **0.91** |

TABLE 3

Investigation of Keywords for Classifying RenD Dataset using BART ZSL Model. Articles are explicitly categorized using both abstract and title. Bold keywords show the best results for that particular category.

| Categories | Keywords | Abstract | | | | | Title | | | | |
|---|---|---|---|---|---|---|---|---|---|---|---|
| | | Ac | F1 | AUC | Pv | Re | Ac | F1 | AUC | Pv | Re |
| **Clinical, Experimental, and Automated Model** | 'Clinical Study', 'Experimental Study', 'Automated Studies' | 0.80 | 0.82 | 0.70 | 0.85 | 0.80 | 0.67 | 0.76 | 0.76 | 0.89 | 0.67 |
| | 'Clinical Study', 'Experimental Study', 'Automated Model', | 0.80 | 0.83 | 0.74 | 0.86 | 0.80 | 0.68 | 0.77 | 0.801 | 0.91 | 0.68 |
| | 'Clinical Study', 'Experimental Study based on animals', 'Technical study based on Automated Model' | 0.85 | 0.87 | 0.91 | 0.92 | 0.85 | 0.85 | 0.87 | 0.91 | 0.92 | 0.85 |
| | **'Clinical Finding based on humans', 'Experimental Study based on animals', 'Technical study based on Automated Model'** | **0.91** | **0.92** | **0.91** | **0.93** | **0.91** | **0.91** | **0.92** | **0.93** | **0.94** | **0.91** |
| **Image processing techniques, Machine learning models, Deep learning models** | 'Deep learning Model', 'Image processing technique', 'ONLY Machine learning' | 0.65 | 0.54 | 0.12 | 0.47 | 0.65 | 0.68 | 0.55 | 0.05 | 0.47 | 0.68 |
| | 'Deep learning Model', 'Image processing technique', 'Classic Machine learning' | 0.66 | 0,57 | 0.74 | 0.79 | 0.66 | 0.69 | 0.60 | 0.73 | 0.71 | 0.69 |
| | 'Deep learning Model', 'Digital Image processing technique', 'Classic Machine learning' | 0.65 | 0.58 | 0.68 | 0.61 | 0.65 | 0.66 | 0.56 | 0.71 | 0.76 | 0.66 |
| | **'Deep learning Model', 'Digital Image processing technique', 'Machine learning Model'** | **0.82** | **0.82** | **0.87** | **0.86** | **0.82** | **0.92** | **0.92** | **0.95** | **0.94** | **0.92** |

multilabel classification scenario. However, ZSL achieved the best score of F1 micro of 0.52, AUC of 0.68, precision micro of 0.49, and recall micro of 0.61. The ocular group is classified into DR, DME, AMD, glaucoma, dry eye, cataract, CSR, and retinal detachment. In terms of accuracy, F1 score, AUC, precision, and recall, the classification based on titles yielded the most favorable outcomes. Specifically, the results for titles-based classification were 0.85 accuracy, 0.85 F1 score, 0.92 AUC, 0.89 precision, and 0.86 recall. Following closely were the results for the abstract-based classification, with values of 0.85 accuracy, 0.83 F1 score, 0.91 AUC, 0.87 precision, and 0.85 recall.

### 3.5.2. Trend Analysis

Category-wise analysis was performed for article type and ocular disease group based on the RenD dataset (Figure 2). Category-wise analysis showed more frequent papers on clinical studies compared to experimental and automated based studies. For ocular diseases, more studies discussed DR and glaucoma compared to other ocular diseases. A time-wise analysis was conducted for the subgroup of automated studies from 2015 to 2022. The trends indicated that in the initial years, these studies primarily relied on image processing techniques. However, as time progressed, machine learning gained traction, and eventually, deep learning models became increasingly popular in this field, reflecting how the technology is evolving in ophthalmology.

## 4. DISCUSSION

We have a proposed framework designed to streamline the literature review process. This framework entails an automated system that operates by taking user-specified keywords as input. Leveraging these keywords, the system retrieves relevant articles from the PubMed database. Additionally, the user specifies the desired categorization for these articles. This approach aims to simplify and expedite the traditionally time-consuming task of conducting literature reviews. We investigated the efficacy of using LLM model to carry out article classification.

LLM models, particularly Chat Generative Pre-training Transformer (ChatGPT), have gained a huge popularity due to their versatility in performing various tasks, including question answering and trend analysis within specific fields. Moreover, these models demonstrate the capability to generate research papers, letters, and other written content, showcasing their potential for creative text generation. However, the generated content is not always entirely authentic, as it can occasionally produce fake references and links, raising concerns about the reliability and accuracy of the information presented. So, we proposed a framework for automating the literature review process and finding different trends in various disciplines. A limitation of ChatGPT 3.5 is however the fact that it is not equipped with information beyond September 2021, therefore, it may not provide facts or knowledge beyond this date. We target to use open source LLMs for article classification and then perform categories-wise and timewise analysis. Other open-source LLMs have become available recently. For instance, Mdeberta, BART, and recently released Llma 2 and its variant that may outperform ChatGPT. However, utilizing Llma 2 requires significant GPU and memory resources. Even the small variant of the model, with 7 billion parameters, demands substantial computational power to function effectively. These resource requirements can pose challenges for users with limited access to high-performance hardware. Therefore, our objective is to choose a model that demands fewer computational resources, making it accessible to a broader range of users. We opted for the BART model, which is open source and can be executed on CPU, offering both good performance and accessibility to a wider user base.

TABLE 4

CLASSIFICATION OF SCIENTIFIC ARTICLES INTO FOUR GROUPS: ARTICLE TYPE, OCULAR DISEASES, CLINICAL STUDIES SUB-CLASS, AND AUTOMATED STUDIES SUB-CLASS WHICH ARE CLASSIFIED INTO 3,8,4, AND 3 CATEGORIES RESPECTIVELY. THE ABERRATIONS ARE, MC: MULTICLASS CLASSIFICATION, AND ML: MULTILABEL CLASSIFICATION. THE BOLD INDICES SHOW THE BEST RESULTS AND BLUE INDICES SHOW SECOND BEST SCORES.

| | | Article Type | Ocular Diseases | Clinical Studies Sub-Class | Automated Studies Sub-Class |
|---|---|---|---|---|---|
| **Classification Type** | | **MC** | **MC** | **ML** | **MC** |
| **Abstract** | Ac | **0.91** | 0.85 | - | 0.82 |
| | F1 | **0.92** | 0.83 | 0.49 | 0.82 |
| | AUC | **0.91** | 0.91 | 0.67 | 0.87 |
| | Pv | **0.93** | 0.87 | 0.33 | 0.86 |
| | Re | **0.91** | 0.85 | 0.82 | 0.82 |
| **Title** | Ac | **0.91** | **0.85** | - | **0.92** |
| | F1 | **0.92** | **0.85** | 0.50 | **0.92** |
| | AUC | **0.91** | **0.92** | 0.67 | **0.95** |
| | Pv | **0.93** | **0.89** | 0.42 | **0.94** |
| | Re | **0.91** | **0.86** | 0.61 | **0.92** |
| **Probability (Abstract +Title)** | Ac | 0.85 | 0.78 | - | 0.90 |
| | F1 | 0.87 | 0.73 | 0.51 | 0.89 |
| | AUC | 0.91 | 0.86 | 0.67 | 0.93 |
| | Pv | 0.92 | 0.73 | 0.42 | 0.91 |
| | Re | 0.85 | 0.78 | 0.61 | 0.90 |
| **Appending Title to Abstract** | Ac | 0.91 | 0.84 | - | 0.90 |
| | F1 | 0.91 | 0.82 | 0.52 | 0.89 |
| | AUC | 0.91 | 0.90 | 0.68 | 0.93 |
| | Pv | 0.93 | 0.86 | 0.49 | 0.91 |
| | Re | 0.91 | 0.84 | 0.61 | 0.90 |

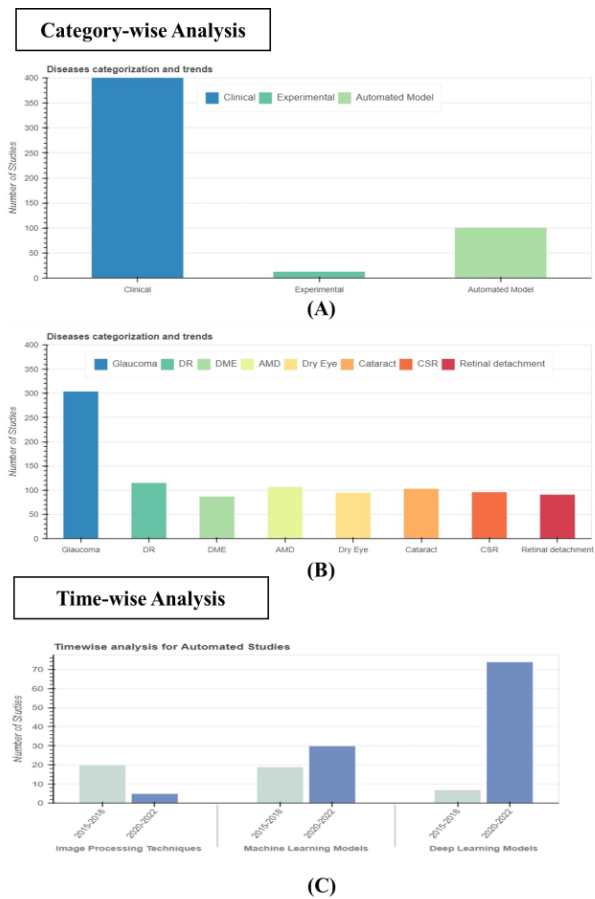

**Category-wise Analysis**

**Time-wise Analysis**

**(A)**

**(B)**

**(C)**

Fig. 2. Trend Analysis of classified articles. (A, B) Categories-wise analysis for article type and ocular diseases group respectively. (C) Timewise analysis for Automated studies subclass group

### 4.1 Categorization Classification and Trend Analysis

We have employed BART as ZSL classifier, we used abstract and title separately for article classification. After obtaining probabilities from each model, we combined the probabilities and performed classification. Additionally, we also appended the title with the abstract and fed into the models.

The BART model showed compromising results for the categories article type, ocular diseases, and automated studies sub-class of RenD dataset. The classification based on abstract, and title are nearly close performance. For clinical studies sub-class grouping, the BART achieved an F1 micro score of 0.52 and an AUC of 0.68. To improve the performance for this class, we fine-tuned BERT and its variant, BioBERT, which performed best with an F1 micro score of 0.67 and an AUC of 0.70. The lower performance in this class is likely due to the class imbalance, that typically affects the model's ability to generalize and accurately predict instances of the minority class.

We also evaluated the performance of BART model to classify the articles into different categories for two undergoing review studies including DEye and DemL. The articles of both review studies were annotated by reviewing the whole article. However, we just used only abstract and title, in which for DEye, our model achieved AUC 0.79 and F1 score 0.63. In order to improve the performance of the BART model, we performed a hierarchical analysis that multilabel task is divided into a binary classification. Classification was carried out across different thresholds for each class, and the optimal value was chosen based on the achieved best results (see supplementary Table 3). Results showed that by converting multilabel problem into binary classification improves the performance of BART model. In addition to this, we also observed that abstract-based classification and whole-article-based annotation provided comparable results for each class (see Table 5). Based on the DemL dataset, the BART model's classification based on abstract, and title is equally accurate as the whole-article-based annotation for the DemL dataset.

### 4.2. Comparative Analysis of Computational Complexity

We have performed a comparative analysis of processing time between the LLM model and human annotators, which has unveiled intriguing insights. This analysis delves into the time required for classification based on abstract, title, and both title and abstract of scientific articles. Manually annotating articles is a time-consuming task as human annotators require significant time to label each article. The overall process can span over several weeks, depending on the number of articles and the number of categories for annotations. For instance, annotating the abstract of one article with two categories may take on average 4 - 5 minutes. However, automated model taking notably less time to complete similar tasks (see Table 6). Notably, utilizing both title and abstract as input led to slightly increased processing time for the BART model, although it remained significantly faster than human annotation.

We conducted two types of trend analysis: category-wise and timewise. These analyses can be applied to any classified category and highlight different trends in a concise and quick manner. A report is generated at the end, encompassing user-specified inclusion criteria and other relevant aspects to aid researchers (see supplementary Figure 1).

These findings emphasize the potential of LLM models in accelerating the literature review process and present a compelling argument for leveraging in research articles, yielding more time-efficient and effective knowledge extraction and categorization. By automating the literature review process, our framework provides researchers with a valuable tool for efficient information retrieval, classification, and trend analysis, facilitating evidence-based decision-making and advancing knowledge in the field of ophthalmology.

As, we have employed the ZSL model for categorization of articles in field of ophthalmology, but it is extendable to other categories and fields without requiring any additional training. The model can generalize to new classes it has not seen during training, making it adaptable to different domains and applications.

The limitation of study is that we have included the articles from the PubMed database, which may result in the exclusion of relevant articles related to the chosen keyword. However, future plans involve the integration of additional databases such as Google Scholar, IEEE Xplore, and Springer to address this limitation and ensure a more comprehensive coverage of relevant literature. Articles from various databases can unveil trends and patterns that transcend specific domains, increasing the applicability of findings.

## 5. Conclusion

We developed a framework based on BART ZSL for categorization and trend analysis of articles and demonstrated a proof-of-concept scenario in the field of ophthalmology. Our model can be used in other biomedical disciplines for text classification. Results demonstrated that the model achieved promising outcomes across most of the categories. In addition to article classification, trend analysis highlighted the evolution of technology in ophthalmology. Accurate and quick classification of scientific papers enables efficient information retrieval, allowing researchers to more quickly access relevant studies, and obtain insights into the trend of technology and future directions. Future research directions include exploring more specialized LLMs for further improvement.

### Acknowledgment

This work was supported by NIH Grants R01EY033005 (SY), R21EY031725 (SY), and Challenge Grant from Research to Prevent Blindness (RPB), New York (SY). The funders had no role in study design, data collection and analysis, decision to publish, or preparation of the manuscript.

### TABLE 5
BART MODEL EVALUATION FOR CLASSIFICATION OF DEYE AND DEML DATASETS. ABBREVIATION THE ABBREVIATIONS ARE: DA: DATASET, TY: CLASSIFICATION TYPE, ML: MULTILABEL CLASSIFICATION, BC: BINARY CLASSIFICATION, MC: MULTICLASS CLASSIFICATION

| $D_a$ | Categories | Ty | Abstract | | | | | Title | | | | |
|---|---|---|---|---|---|---|---|---|---|---|---|---|
| | | | Ac | F1 | AUC | Pv | Re | Ac | F1 | AUC | Pv | Re |
| **DEye** | Tear Film Break Up Time, Infrared Thermography, Lipid Layer Interface Pattern, Meibomian Gland Study, Tear Film Assessment, Tear Meniscus Assessment | MC | 0.63 | 0.79 | 0.60 | 0.67 | 0.63 | 0.44 | 0.72 | 0.28 | 0.84 | 0.44 |
| | **Tear Film Break Up Time** | BC | **0.91** | **0.73** | **0.79** | **1.0** | **0.58** | **0.7** | **0.47** | **0.72** | **0.34** | **0.75** |
| | Infrared Thermography | BC | 0.94 | 0.77 | 0.91 | 0.70 | 0.89 | 0.83 | 0.42 | 0.69 | 0.39 | 0.5 |
| | Lipid Layer Interface Pattern | BC | 0.91 | 0.54 | 0.71 | 0.80 | 0.44 | 0.86 | 0.47 | 0.68 | 0.50 | 0.44 |
| | Meibomian Gland Study | BC | 0.92 | 0.87 | 0.90 | 0.89 | 0.85 | 0.92 | 0.87 | 0.90 | 0.89 | 0.85 |
| | Tear Film Assessment | BC | 0.80 | 0.64 | 0.80 | 0.54 | 0.80 | 0.71 | 0.42 | 0.62 | 0.38 | 0.46 |
| | Tear Meniscus Assessment | BC | **0.98** | **0.85** | **0.87** | **1.0** | **0.75** | **1.0** | **1.0** | **1.0** | **1.0** | **1.0** |
| **DemL** | Machine learning model, Deep learning model | MC | **0.99** | **0.99** | **0.98** | **0.99** | **0.99** | **0.99** | **0.99** | **0.98** | **0.99** | **0.99** |

### TABLE 6
COMPARATIVE ANALYSIS OF PROCESSING TIME BY LLM MODEL (BART), AND HUMAN ANNOTATOR. TIME CALCULATED FOR CLASSIFICATION BASED ON ABSTRACT, TITLE AND TITLE & ABSTRACT

| Dataset | Categories | Articles | Abstract (min) | Title (min) | Title & Abstract (min) | Annotation by Human (min) | Timeline (months) |
|---|---|---|---|---|---|---|---|
| **RenD** | Clinical, Experimental, and Automated Model | 1000 | 141.50 | 15.21 | 194.2 | ~3000 | 4 |
| | DR, DME, AMD, glaucoma, Dry Eye, Cataract, CSR, and retinal detachment | 1000 | 274.06 | 27.30 | 283.43 | ~4000 | |
| | Screening, Diagnosis, Prognosis, Etiology, and Management | 464 | 118.71 | 13.11 | 120.34 | ~1600 | 4 |
| | Image processing techniques, Machine learning model, and Deep learning model | 156 | 21.23 | 2.45 | 23.78 | ~400 | 4 |
| **DEye** | Tear Film Break Up Time, Infrared Thermography, Lipid Layer Interface Pattern, Meibomian Gland Study, Tear | 67 | 36.45 | 6.45 | 37.12 | ~2800 | 2 |
| | Film Assessment, Tear Meniscus Assessment | | | | | | |
| **DemL** | Deep learning Model, Machine learning Model | 115 | 19.30 | 1.83 | 32.23 | ~1000 | 1 |

## References

[1] Á. O. Dos Santos, E. S. da Silva, L. M. Couto, G. V. L. Reis, and V. S. Belo, "The use of artificial intelligence for automating or semi-automating biomedical literature analyses: a scoping review," *Journal of Biomedical Informatics*, p. 104389, 2023.

[2] H. Kilicoglu, D. Demner-Fushman, T. C. Rindflesch, N. L. Wilczynski, and R. B. Haynes, "Towards automatic recognition of scientifically rigorous clinical research evidence," *Journal of the American Medical Informatics Association*, vol. 16, no. 1, pp. 25-31, 2009.

[3] C. Lokker *et al.*, "Machine learning to increase the efficiency of a literature surveillance system: a performance evaluation," *medRxiv*, p. 2023.06. 18.23291567, 2023.

[4] K. Hashimoto, G. Kontonatsios, M. Miwa, and S. Ananiadou, "Topic detection using paragraph vectors to support active learning in systematic reviews," *Journal of biomedical informatics*, vol. 62, pp. 59-65, 2016.

[5] M. M. Kebede, C. Le Cornet, and R. T. Fortner, "In-depth evaluation of machine learning methods for semi-automating article screening in a systematic review of mechanistic literature," *Research Synthesis Methods*, vol. 14, no. 2, pp. 156-172, 2023.

[6] A. Kanegasaki, A. Shoji, K. Iwasaki, and K. Kokubo, "PRM75-DEVELOPMENT OF MACHINE LEARNING BASED ABSTRACT DOCUMENT CLASSIFICATION FOR SUPPORTING SYSTEMATIC REVIEWS," *Value in Health*, vol. 21, pp. S368, 2018.

[7] S. Y. Wang, J. Huang, H. Hwang, W. Hu, S. Tao, and T. Hernandez-Boussard, "Leveraging weak supervision to perform named entity recognition in electronic health records progress notes to identify the ophthalmology exam," *International Journal of Medical Informatics*, vol. 167, p. 104864, 2022.

[8] A. N. Yew, M. Schraagen, W. M. Otte, and E. van Diessen, "Transforming epilepsy research: A systematic review on natural language processing applications," *Epilepsia*, vol. 64, no. 2, pp. 292-305, 2023.

[9] C. Li, Y. Zhang, Y. Weng, B. Wang, and Z. Li, "Natural language processing applications for computer-aided diagnosis in oncology," *Diagnostics*, vol. 13, no. 2, p. 286, 2023.

[10] W. Hariri, "Unlocking the Potential of ChatGPT: A Comprehensive Exploration of its Applications, Advantages, Limitations, and Future Directions in Natural Language Processing," *arXiv preprint arXiv:2304.02017*, 2023.

[11] I. Domocos *et al.*, "BERT for Complex Systematic Review Screening to Support the Future of Medical Research," in *Artificial Intelligence in Medicine: 21st International Conference on Artificial Intelligence in Medicine, AIME 2023, Portorož, Slovenia, June 12–15, 2023, Proceedings*: Springer Nature, p. 173.

[12] A. K. Ambalavanan and M. V. Devarakonda, "Using the contextual language model BERT for multi-criteria classification of scientific articles," *Journal of biomedical informatics*, vol. 112, p. 103578, 2020.

[13] J. Devlin, M.-W. Chang, K. Lee, and K. Toutanova, "Bert: Pre-training of deep bidirectional transformers for language understanding," *arXiv preprint arXiv:1810.04805*, 2018.

[14] C. Lokker *et al.*, "Deep learning to refine the identification of high-quality clinical research articles from the biomedical literature: Performance evaluation," *Journal of Biomedical Informatics*, vol. 142, p. 104384, 2023.

[15] N. Mylonas, S. Karlos, and G. Tsoumakas, "WeakMeSH: Leveraging provenance information for weakly supervised classification of biomedical articles with emerging MeSH descriptors," *Artificial Intelligence in Medicine*, vol. 137, p. 102505, 2023.

[16] Y.-S. Wang, T.-C. Chi, R. Zhang, and Y. Yang, "PESCO: Prompt-enhanced Self Contrastive Learning for Zero-shot Text Classification," *arXiv preprint arXiv:2305.14963*, 2023.

[17] L. Gao, D. Ghosh, and K. Gimpel, "The Benefits of Label-Description Training for Zero-Shot Text Classification," *arXiv preprint arXiv:2305.02239*, 2023.

[18] M. Pàmies *et al.*, "A weakly supervised textual entailment approach to zero-shot text classification," in *Proceedings of the 17th Conference of the European Chapter of the Association for Computational Linguistics*, 2023, pp. 286-296.

[19] Y. Zhang *et al.*, "Metadata-induced contrastive learning for zero-shot multi-label text classification," in *Proceedings of the ACM Web Conference 2022*, 2022, pp. 3162-3173.

[20] N. Mylonas, S. Karlos, and G. Tsoumakas, "Zero-shot classification of biomedical articles with emerging mesh descriptors," in *11th hellenic conference on artificial intelligence*, 2020, pp. 175-184.

[21] N. L. Wilczynski, D. Morgan, R. B. Haynes, and H. T. w. m. ca, "An overview of the design and methods for retrieving high-quality studies for clinical care," *BMC medical informatics and decision making*, vol. 5, pp. 1-8, 2005.

[22]   M. Lewis *et al.*, "Bart: Denoising sequence-to-sequence pre-training for natural language generation, translation, and comprehension," *arXiv preprint arXiv:1910.13461,* 2019.